\documentclass[11pt,a4paper]{article}
\usepackage[hyperref]{emnlp2018}
\usepackage{times}
\usepackage{latexsym}

\usepackage{url}

\usepackage{amsmath}
\usepackage{adjustbox}
\usepackage{booktabs}
\usepackage{pgfplots}
\usepackage{pgfplotstable}
\usepackage{multirow}
\usepackage{caption}

\pgfplotsset{compat=newest}
\pgfplotsset{every mark/.append style={solid}}

\aclfinalcopy 

\title{
    Wronging a Right:
    Generating Better Errors to Improve \\ Grammatical Error Detection
}

\author{Sudhanshu Kasewa  \and Pontus Stenetorp \and Sebastian Riedel \\
        \{sudhanshu.kasewa.16, p.stenetorp, s.riedel\}@ucl.ac.uk \\
        University College London}

\date{}

\begin{document}
\maketitle

\begin{abstract}
    Grammatical error correction, like other machine learning tasks, greatly benefits from large quantities of high quality training data, which is typically expensive to produce.
    While writing a program to automatically generate realistic grammatical errors would be difficult, one could learn the distribution of naturally-occurring errors and attempt to introduce them into other datasets.
    Initial work on inducing errors in this way using statistical machine translation has shown promise; we investigate cheaply constructing synthetic samples, given a small corpus of human-annotated data, using an off-the-rack attentive sequence-to-sequence model and a straight-forward post-processing procedure.
    Our approach yields error-filled artificial data that helps a vanilla bi-directional LSTM to outperform the previous state of the art at grammatical error detection, and a previously introduced model to gain further improvements of over 5\% $F_{0.5}$ score.
    When attempting to determine if a given sentence is synthetic, a human annotator at best achieves 39.39 $F_{1}$ score, indicating that our model generates mostly human-like instances.
\end{abstract}

\section{Introduction}
There is an ever-growing number of people learning English as a second language; providing them with quick feedback to facilitate their learning is a crucial, labour-intensive endeavour.
Part of this process is identifying and correcting grammatical errors, and several computational techniques have been developed to automate it~\citep{rozovskaya2014building,junczysdowmunt2016phrase}.
For example, given an erroneous sentence \textit{``I wanted to \textbf{goes} to the beach''}, the grammatical error correction task is to output the valid sentence \textit{``I wanted to \textbf{go} to the beach''}.
The task can be cast as a two-stage process, detection and correction, which can either be performed sequentially~\citep{yannakoudakis2017neural}, or jointly~\citep{napoles2017systematically}.

Automated error correction performance is arguably still too low for practical consideration, perhaps limited by the amount of training data~\citep{rei2017artificial}.
High quality annotations are expensive to procure, and foreign language learners and commercial entities may feel uncomfortable granting access to their data.
Instead, one could attempt to supplement existing manual annotations with synthetic instances.
%
Such artificial samples are beneficial only when they share structure with the true distribution from which human errors are generated.
Generative Adversarial Networks~\citep{goodfellow2014generative} could be used for this purpose, but they are difficult to train, and require a large collection of sentences that are incorrect. 
One might attempt 
self-training~\citep{mcclosky2006effective}, where new instances are generated by applying a trained model to unannotated data, using high-confidence predictions as ground truth labels.
However, in such a scheme, the expectation is that the unlabelled text already contains errors, which is not usually the case for most freely available text such as Wikipedia articles as they strive towards correctness.
%

In place of using machine translation~(MT) to \textit{correct} grammatical mistakes~\citep{yuan2013constrained,junczysdowmunt2014theamu,yuan2016grammatical}, one might consider swapping the input and output streams, and instead learn to \textit{induce errors} into error-free text, for the purpose of creating a synthetic training dataset~\citep{felice2014generating}. 
Recently, \citet{rei2017artificial} used a statistical MT~(SMT) system to induce errors into error-free text.
Building on this work, and leveraging recent advances in neural MT~(NMT), we used an off-the-shelf attentive sequence-to-sequence model~\citep{britz2017massive}, eliminating the need of specialised software such as a phrase-table generator, decoder, and part-of-speech tagger.
We created multiple synthetic datasets from in-domain and out-of-domain sources, and found that stochastic token sampling, and pruning redundant and low-likelihood sentences, were helpful in generating meaningful corruptions.
Using the artificial samples thus generated, we improved upon detection results with simply a vanilla bi-directional LSTM~\citep{hochreiter1997lstm}.
Using a more powerful model, we established new state-of-the-art results, that improve on previously published $F_{0.5}$ scores by over 5\%.
Additionally, we confirm that our generated instances are human-like, as an annotator identifying generated sentences achieved a maximum $F_{1}$ score of 39.39.

\section{Related work}
In computer vision, images are blurred, rotated, or otherwise deformed inexpensively to create new training instances~\citep{wang2017effectiveness}, because such manipulation does not significantly alter the image semantics.
Similar coarse processes do not work in NLP since mutating even a single letter or a word can change a sentence's meaning, or render it nonsensical.
Nonetheless, \citet{vinyals2015grammar} employed a kind of self-training where they use noisy predictions for unlabelled instances output by existing state-of-the-art parsers as ground-truth labels, and improved syntactic parsing performance.
\citet{sennrich2016improving} synthesised training instances by round-trip-translating a monolingual corpus with weaker versions of an NMT learner, and used them to improve the 
translation.
\citet{bouchard2016learning} developed an efficient algorithm to blend generated and true data for improving generalisation.

Grammar correction is a well-studied task in NLP, and early  systems were rule-based pattern recognisers~\citep{macdonald1983human} and dictionary-based linguistic analysis engines~\citep{richardson1988experience}.
Later systems used statistical approaches, addressing specific kinds of errors such as article insertion~\citep{knight1994integrating} and spelling correction~\citep{golding1996applying}.
Most recently, architectural innovations in neural sequence labelling~\citep{rei2016attending,rei2017semi} raised error detection performance through improved ability to process unknown words and jointly learning a language model.

Early efforts for artificial error generation included generating specific types of errors, such as mass noun errors~\citep{brockett2006correcting} and article errors~\citep{rozovskaya2010training}, and leveraging linguistic information to identify error patterns and transfer them onto grammatically correct text~\citep{foster2009generrate,yuan2013constrained}.
\citet{imamura2012grammar} investigated methods to generate pseudo-erroneous sentences for error correction in Japanese. 
Recently, \citet{rei2017artificial} corrupted error-free text using SMT to create training instances for error detection.

\begin{table*}
\small
\centering
\resizebox{0.7\linewidth}{!}{%
    \begin{tabular}{ll} 
        Original & Corruption \\ \midrule
        She promised to turn over a new leaf. & She \underline{promissed} to turn over a new leaf.  \\
        At the moment I'm in Spain. & \underline{During} the moment I'm in Spain.  \\
    \end{tabular}%
}
\caption{
    Example sentences generated by our NMT pipeline.
}
\label{tbl:examples}
\end{table*}
\begin{table*}
\small
\centering
\resizebox{1\linewidth}{!}{%
\begin{tabular}{llrrrr}
    Data augmentation strategy & Model & FCE (dev) & FCE & CoNLL1 & CoNLL2 \\
    \midrule
    \citet{rei2017artificial} FCE$_{PAT}$ + EVP$_{PAT}$ & SL    & --    & 47.8 & 19.5 & 28.5 \\
    \citet{rei2017artificial} FCE$_{SMT}$ + EVP$_{SMT}$ & SL    & --    & 48.4 & 19.7 & 28.4 \\
    \citet{rei2017artificial} FCE$_{SMT+PAT}$ + EVP$_{SMT+PAT}$ & SL    & --    & 49.1 & 21.9 & 30.1 \\
    \midrule
    None                        & BiLSTM  & 47.9 & 43.6 & 16.6 & 24.3 \\
    FCE$_{TS}$                   & BiLSTM  & 51.2 & 47.1 & 19.7 & 28.9 \\
    EVP$_{BS}$                   & BiLSTM  & 52.1 & 50.1 & 20.8 & 29.0 \\
    SW$_{TS}$                    & BiLSTM  & 51.5 & 50.6 & 24.2 & 31.7 \\
    FCE$_{AM+TS}$+EVP$_{AM+TS}$         & BiLSTM  & 52.3 & 50.4 & 22.1 & 30.8 \\
    \midrule
    None                        &SL    & 52.5 & 48.2 & 17.4 & 25.5 \\
    FCE$_{TS}$                   & SL    & 54.8 & 49.9 & 20.9 & 29.2 \\
    EVP$_{BS}$                   & SL    & 55.2 & 54.6 & 23.3 & 31.4 \\
    SW$_{TS}$                    & SL    & 53.8 & 52.7 & 26.8 & 34.3 \\
    FCE$_{AM+TS}$+EVP$_{AM+TS}$ & SL    & \textbf{56.9} &  54.6 & 25.1 &  33.0 \\
    FCE$_{AM+TS}$+EVP$_{AM+TS}$+SW$_{AM+TS}$   & SL    & 56.5 &  \textbf{55.6}    &  \textbf{28.3}    & \textbf{35.5} \\
    \bottomrule
\end{tabular}
}
\caption{
    $F_{0.5}$ scores on various tests contrasted with published results and unaugmented baseline models.
}
\label{tbl:results}
\end{table*}

\section{Neural error generation}
To learn to introduce errors, we use an off-the-shelf attentive sequence-to-sequence neural network~\citep{bahdanau2014neural}.
Given an input sequence, the \textit{encoder} generates context vectors for each token.  
%
%
Then, the \textit{attention mechanism} and the \textit{decoder} work in tandem to emit 
a distribution over the target vocabulary.
At every decoder time-step, the encoder context vectors are scored by the attention mechanism, and a weighted sum is supplied to the decoder, along with its propagated internal state and last output symbol.
%
%

\paragraph{\textbf{Corruption:}}
Tokens from this distribution are sampled at every decoder time-step, either by $argmax$~(AM), which emits the most likely word, or by a stochastic alternative such as \textit{temperature sampling}~(TS) as $argmax$ cannot be relied on to generate rare words.
A temperature parameter $\tau > 0$ sharpens or softens the distribution:
\begin{align*}
    \tilde{p}_i = \textit{\textbf{f}}_{\tau}(p)_i = \dfrac{p_i^{\frac{1}{\tau}}}{\sum_j p_j^{\frac{1}{\tau}}}
\end{align*}
where $i$ are the components of the probability distribution corresponding to words in the vocabulary.
As one interpolates $\tau$ from 0 to 1, the behaviour of $\tilde{p}$ transitions from \textit{argmax} to $p$, controlling the diversity of the generated tokens.

The sentence generated by TS might be a low probability sequence from the joint conditional distribution $P(\textbf{v} | \textbf{u}) $, where $\textbf{u}$ is the input sentence and $\textbf{v}$ is the output sentence.
One way around this is to use \textit{beam search}~(BS), which checks the likelihood of every possible continuation of a sentence fragment, and maintains a list of the $n$ best translations generated up to the current time-step.
AM, TS, and BS are indicative of the trade-off between increasing levels of model flexibility at the cost of computation; we compare them to assess whether the additional computations were helpful in creating high-quality synthetic instances.

\paragraph{\textbf{Post-processing:}}
Original and corrupted sentences are aligned at a word-level using Levenshtein distance.
%
%
Using the minimal alignment, words in the corrupted sentence are labelled \textit{correct}, `c', or \textit{incorrect}, `i', as follows:

\textit{If} the word is not aligned with itself, then `i'.
\textit{Else}, if following a gap, then `i', as at this point a human reader would notice that there is a word missing in the sentence.
\textit{Else}, if it is the last word, but it is not aligned to the last word of the source sentence, then `i', as a human would realise that this sentence ends abruptly,
\textit{Else}, `c'.

These token-labelled corrupted sentences now form an artificial dataset for training an error detector.
Duplicate instances and corrupted sentences with more than 5 errors were dropped to remove noise from the downstream training.

\section{Experiments}
We evaluated our approach on the First Certificate of English~(FCE) error detection dataset~\citep{rei2016compositional}, as well as on two human-annotated test sets~(CoNLL1, CoNLL2) from the CoNLL~2014 shared task~\citep{ng2014conll}.
The CoNLL data sets pose a unique challenge; as they are different in style and domain from FCE, we have no matching training data.
We compared the effect of different neural generation procedures~(AM, TS, BS) and contrasted the downstream performance of a bidirectional LSTM with an elaborate sequence labeller.
%
%

\subsection{Implementation details}
\paragraph{\textbf{NMT training and corruption:}}
We minimally modified the open source implementation%
\footnote{
    \url{https://github.com/google/seq2seq}
} of \citet{britz2017massive} to implement TS and BS.%
\footnote{
    \url{https://github.com/skasewa/wronging}
}
We trained our NMT with a single-layered encoder and decoder with cell size 256, on the parallel corpus version of FCE~\citep{yannakoudakis2011anew}, with early stopping after the FCE development set score dropped consistently for 20 epochs.
We introduced errors into three datasets: FCE itself (450K tokens), the English Vocabulary Profile or EVP (270K tokens) and a subset of Simple Wikipedia or SW (8.4M tokens); of these, FCE and EVP were both used in artificial error generation via SMT and pattern extraction (PAT) by \citet{rei2017artificial}, enabling us to make a fair experimental comparison.
Ten corrupted versions using each of AM, TS~($\tau = 0.05$) and BS were sampled for FCE and EVP corruptions, while one sufficed for SW.
The theoretical time complexity of BS is O($bn$) for each sentence, where $b$ is number of candidates, and $n$ is the maximum length of a sentence.
Empirically, BS with $b=11$ took a factor of 11.3 more time than AM.
Examples of generated errors are provided in Table~\ref{tbl:examples}.

\paragraph{\textbf{Error detection:}}
We compare two error detection models: a vanilla bi-directional LSTM~(BiLSTM)~\citep{schuster1997bidirectional}, and the state-of-the-art sequence labeller~(SL) neural network used by \citet{rei2017artificial}.
These models were trained on the binary-labelled FCE training set augmented with the corrupted instances.
Wherever no model is explicitly
stated, the SL model was used.
%
During training, we alternate between the annotated FCE dataset and the synthetic collection.
This alternating protocol prevents overfitting on 
FCE; 
once it shifts back, it reinforces connections made from the helpful synthetic corruptions while forgetting about the noisy ones.

\subsection{Results}
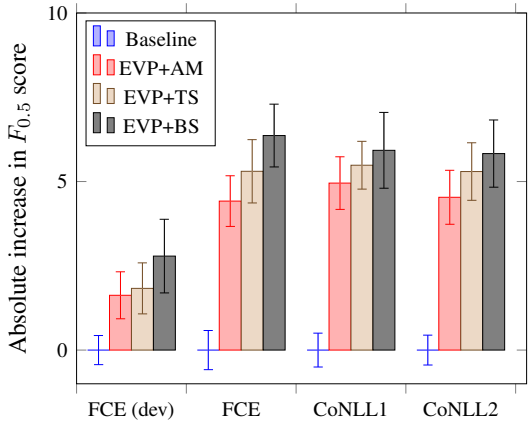
\begin{figure}
\begin{adjustbox}{width=0.9\columnwidth}
\begin{tikzpicture}
\begin{axis}[
ybar=0pt,
bar width=0.2,
xtick={0.5,1.5,...,5.5},
xticklabels={FCE (dev),FCE, CoNLL1, CoNLL2},
ytick={0,0.05,0.1},
yticklabels={0, 5, 10},
tick label style={font=\small},
legend style={font=\small},
ylabel=Absolute increase in $F_{0.5}$ score,
x tick label as interval,
xmin=0.5,xmax=4.6,
ymin=-0.01,ymax=0.1,
legend style={at={(0.02,0.98)},anchor=north west}
]

\addplot+[error bars/.cd,
y dir=both,y explicit]
coordinates {
(1,0) +- (0.0,0.00431888)
(2,0) +- (0.0,0.00580828)
(3,0) +- (0.0,0.00502179)
(4,0) +- (0.0,0.00442995)
    };
\addplot+[error bars/.cd,
y dir=both,y explicit]
coordinates {
(1,0.01625371) +- (0.0,0.00696896)
(2,0.04419627) +- (0.0,0.00751478)
(3,0.04953643) +- (0.0,0.00780515)
(4,0.04532661) +- (0.0,0.00800651)
    };
\addplot+[error bars/.cd,
y dir=both,y explicit]
coordinates {
(1,0.01830529) +- (0.0,0.00756621)
(2,0.05303467) +- (0.0,0.00939051)
(3,0.05483744) +- (0.0,0.00708804)
(4,0.05296372) +- (0.0,0.00853338)
    };
\addplot+[error bars/.cd,
y dir=both,y explicit]
coordinates {
(1,0.02788075) +- (0.0,0.01092295)
(2,0.06364042) +- (0.0,0.00931307)
(3,0.05925818) +- (0.0,0.01124607)
(4,0.0582898) +- (0.0,0.00997017)
    };
\legend{Baseline, EVP+AM, EVP+TS, EVP+BS}
\end{axis}
\end{tikzpicture}
\end{adjustbox}
\caption{
    Improvements using three different methods of generation.
}
\label{fig:sampling}
\end{figure}

\begin{figure}
\centering
\begin{adjustbox}{width=0.85\columnwidth}
\begin{tikzpicture}
\begin{axis}[name=boundary,
xtick={1,...,11},
xticklabels={1x, 2x, 3x, 4x, 5x, 6x, 7x, 8x, 9x, 10x},
tick label style={font=\small},
legend style={font=\small},
xlabel=Amount of augmented data,
tick label style={font=\small},
legend style={font=\small},
xmin=0,xmax=11,
ymin=0.15,ymax=0.35,
ylabel=$F_{0.5}$ score,
ytick={0.15, 0.2,0.25,0.3,0.35},
yticklabels={15, 20,25,30,35},
legend style={at={(0.98,0.02)},anchor=south east}
]
\addplot[color=blue] table[x=X, y=T3]{fce.dat};\label{pgfplots:c1r1}
\addplot[color=red] table[x=X, y=T4]{fce.dat};\label{pgfplots:c1r2}
\addplot[color=blue,mark=triangle*] table[x=X, y=T3]{sw.dat};\label{pgfplots:c2r1}
\addplot[color=red,mark=triangle*] table[x=X, y=T4]{sw.dat};\label{pgfplots:c2r2}
\end{axis}
  \node[draw,fill=white,inner sep=1pt,above left=0.5em] at (boundary.south east) {\small
    \begin{tabular}{ccl}
    FCE & SW \\
    \ref{pgfplots:c1r1} & \ref{pgfplots:c2r1} & CoNLL1\\
    \ref{pgfplots:c1r2} & \ref{pgfplots:c2r2} & CoNLL2\\
    \end{tabular}};
\end{tikzpicture}
\end{adjustbox}
\caption{
    Training with increasing amounts of corrupted data from FCE and SW.
}
\label{fig:conll}
\end{figure}
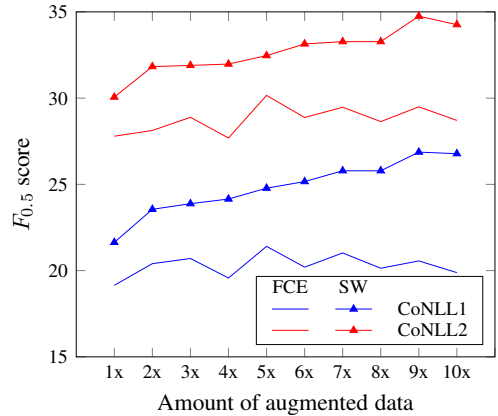

The results for our baselines and data augmentation strategies can be found in Table~\ref{tbl:results}.
Augmented with our NMT generated data, even our vanilla downstream BiLSTM outperforms the SMT+PAT artificial error augmentation approach of~\citet{rei2017artificial}, indicating that our process better generalises the error information in the source dataset.
Using the more powerful SL network bests the previous state of the art by over 5\% on the FCE test. 
Most intriguingly, we note a significant improvement for the CoNLL tests using corruptions from out-of-domain SW.
Figure~\ref{fig:conll} illustrates how we gain performance on these tests with increasing amounts of corrupted SW, which does not hold true for corrupted FCE.
This shows that we were able to induce useful errors into a corpus with a large unseen vocabulary and different syntactic biases, and this in turn proved valuable for detecting errors in a third domain, suggesting that our method can transfer learned distributions across stylistic genres.

Using EVP as a standard source, Figure~\ref{fig:sampling} illustrates the variance of the different sampling methods.
All generation methods yield corruptions that significantly improve test performance, with instances sampled by beam-search consistently outperforming the alternatives.
%

%

\section{Discussion}

\subsection{Error distribution}

The original FCE dataset was annotated using the error taxonomy specified in \citet{nicholls2003cambridge}, and contains 75 unique error codes.
We annotated samples of EVP corrupted by all three sampling methods, at a reduced resolution, to compare the distribution of errors across FCE and the synthetic corpora.
These are presented in Table~\ref{tab:errordist}.

At a high level, NMT generates errors more often among more common parts-of-speech, favouring errors in verbs and nouns, rather than in adverbs and conjunctions. 
It did not make spelling errors as often as in the source dataset; this is likely because it only observed the specific spelling errors present in FCE, and as the vocabulary is restricted to that dataset, it does not encounter those words as frequently in EVP, and thus rarely makes the same spelling mistakes.

Additionally, the differences in these distributions can partially be attributed to the implicit differences between us and the annotators of FCE.

\begin{table}[!bt]
\begin{center}
\resizebox{0.8\linewidth}{!}{%
\begin{tabular}{lrrrr}
\textbf{Spelling} & FCE & AM & TS & BS \\
\midrule
Spelling errors                  & 11  & 1  & 1  & 4  \\
\bottomrule
                                    &     &    &    &    \\
\textbf{Part-of-speech} & FCE & AM & TS & BS \\
\midrule
Verb                                & 34  & 16 & 26 & 16 \\
Preposition                         & 18  & 16 & 10 & 14 \\
Determiner                          & 16  & 7  & 6  & 10 \\
Noun                                & 13  & 36 & 35 & 43 \\
Pronoun                             & 7   & 3  & 3  & 1  \\
Adverb                              & 5   & 5  & 3  & 5  \\
Adjective                           & 3   & 15 & 16 & 12 \\
Conjunction                         & 2   & 2  & 2  & 1  \\
Quantifier                          & 1   & 0  & 0  & 0  \\
\bottomrule
                                    &     &    &    &    \\
\textbf{Remedy Type}           & FCE & AM & TS & BS \\
\midrule
Replacement                         & 49  & 35 & 34 & 32 \\
Inclusion                           & 23  & 30 & 27 & 35 \\
Removal                           & 14  & 33 & 36 & 32 \\
Word form                           & 9   & 2  & 2  & 1  \\
Word order                          & 5   & 0  & 0  & 0 \\
\bottomrule
\end{tabular}
}
\caption{
    Error distribution across FCE and manually annotated samples of artificial data. Spelling errors are a \% of all errors, while Part-of-speech, and Remedy Type are compared within their own categories to sum to 100\%.
}
\label{tab:errordist}
\end{center}
\end{table}

\subsection{Comparison with human errors}

To check if the synthetic instances passed for human-like, we mixed 50 generated sentences among an equal number of actual ungrammatical instances from FCE-dev and tasked a human evaluator to identify the artificial statements, in a simple Turing-style test.
We created three such sets, one for each of our sampling techniques, and the test subject aimed to identify synthetic samples with high confidence.
Results of this test are presented in Table \ref{tab:turing}.

The high precision but low recall scores suggest that while it is still possible to spot some corruptions that are quite clearly artificial, the bulk of our samples do not betray their synthetic nature and are indistinguishable from naturally occurring erroneous sentences.
In order to fairly compare our work with earlier results, we intended to conduct such a test for sentences generated by the SMT of \citet{rei2017artificial}. Unfortunately, we were only able to source corruptions of FCE-train via this method; therefore, we decided not to perform this test as its results cannot be compared to ours.

\begin{table}[!t]
\begin{center}
\resizebox{0.7\linewidth}{!}{%
\begin{tabular}{lrrr}
            & AM   & TS & BS   \\
            \midrule
Precision   & 81.25 & 63.63 & 50.00  \\
Recall      & 26.00 & 28.00 & 14.00 \\
F1          & 39.39 & 38.89 & 22.22 \\
\bottomrule
\end{tabular}
}
\caption{
    Results of a Turing-style test, where a subject was asked to distinguish between real and fake sentences, sampled from each of the different generated corpora.
}
\label{tab:turing}    
\end{center}
\end{table}

\section{Conclusions and future work}
We presented a novel data augmentation technique for grammatical error detection using neural machine translation to learn the distribution of language-learner errors, and induce such errors into grammatically correct text.
We explored several different variants of sampling to improve the quality of our synthetic errors.
After creating artificial training instances with an off-the-shelf NMT, we bettered previous state-of-the-art results on the canonical test with even a basic BiLSTM, and established a new state of the art using a stronger model.
Additionally, we demonstrated that we were able to leverage corruptions of an out-of-domain dataset to set new benchmarks on separate, also out-of-domain tests, without specifically optimising for either.

Our work indicates that neural error generation warrants further investigation with different datasets and architectures, both for error detection and error correction.
Among possible future work is using generative adversarial networks as corruption engines, and developing better sequence alignment methods.
Some preliminary results with simple corruptions using word substitution and word dropout \citep{iyyer2015deep} appear to be promising, and may feature as components of a future corruption system.
Finally, one could use such artificial error-prone corpora as source text for self-training an error detection system.

\ifaclfinal
\section*{Acknowledgements}
We thank Marek Rei and Mariano Felice for granting access to their data and code.
Also, we would like to thank the anonymous reviewers and Johannes Welbl for valuable feedback and discussions.
This work was supported by an Allen Distinguished Investigator Award.
\fi

\bibliographystyle{acl_natbib_nourl}
\bibliography{emnlp2018}

\end{document}